\documentclass[conference]{IEEEtran}

\ifCLASSINFOpdf

\else

\fi

\usepackage[cmex10]{amsmath}
\usepackage{amsthm}
\usepackage{amssymb}
\usepackage{mathrsfs}
\usepackage{graphicx}
\usepackage{float}
\usepackage{array}
\usepackage{epstopdf}
\usepackage{multirow}

\usepackage{times}
\usepackage{epsfig}
\usepackage{amsmath}
\usepackage{multirow}
\usepackage{blkarray}
\usepackage{mathtools}
\usepackage{subcaption}
\usepackage[T1]{fontenc}

\begin{document}
\title{A Robust Regression Approach for Background/Foreground Segmentation}

\author{\IEEEauthorblockN{Shervin Minaee}
\IEEEauthorblockA{ECE Department\\ 
New York University\\
shervin.minaee@nyu.edu\\}
\and
\IEEEauthorblockN{Haoping Yu}
\IEEEauthorblockA{Media Lab\\ 
Huawei Technologies\\
haoping.yu@huawei.com\\}
\and
\IEEEauthorblockN{Yao Wang}
\IEEEauthorblockA{ECE Department\\ 
New York University\\
yw523@nyu.edu\\}
}

\maketitle

\begin{abstract}
Background/foreground segmentation has a lot of applications in image and video processing. In this paper, a segmentation algorithm is proposed which is mainly designed for text and line extraction in screen content. The proposed method makes use of the fact that the background in each block is usually smoothly varying and can be modeled well by a linear combination of a few smoothly varying basis functions, while the foreground text and graphics create sharp discontinuity. The algorithm separates the background and foreground pixels by trying to fit pixel values in the block into a smooth function using a robust regression method. The inlier pixels that can fit well  will be considered  as background, while remaining outlier pixels will be considered foreground. This algorithm has been extensively tested on several images from HEVC standard test sequences for screen content coding, and is shown to have superior performance over other methods, such as the k-means clustering based segmentation algorithm in DjVu. This background/foreground segmentation can be used in different applications such as: text extraction, separate coding of background and foreground for compression of screen content and mixed content documents, principle line extraction from palmprint and crease detection in fingerprint images.
\end{abstract}


\IEEEpeerreviewmaketitle
\IEEEoverridecommandlockouts

\section{Introduction}
Compression and transmission of screen content (i.e. images of the computer/phone screen) is becoming increasingly important to enable online collaboration, online gaming and virtual desktop applications, to name a few.  The screen content contains a lot of text and graphics making them different from photographic images \cite{SCC}. The usual transform based compression methods such as JPEG2000 \cite{jpeg2000} would not necessarily be efficient for compressing and transmitting this category of images. The reason is that, there are a lot of sharp edges in these images which will lead to many non-zero transform coefficients and consequently require a high bit rate. 
A similar problem is compression of scanned color documents, which often consist of mixed contents in the same page, with both smoothly varying background and text and line graphics overlay. \cite{djvu}.
One effective way to compresses such images (screen content or mixed document images) is to decompose the image into three layers; the background layer contains the smooth part of the image, the foreground layer contains text and graphics, and finally a binary mask layer which indicates which layer each pixel belongs to.
The first layer can be coded with transform based coding while the foreground layer can be coded with other techniques designed to exploit the characteristics of text and graphics.  The mask layer can be coded using a binary image coding method.
A key to the success of such coding methods is to accurately separate foreground and background.  The segmentation method proposed here is motivated for the compression of screen content and scanned documents. However, it is not limited to such applications. Indeed, we will show the proposed method can also be used to extract the principle lines in palmprint images as well.

In this paper we look at background/foreground segmentation from a robust regression \cite{robustregression} point of view which to the best of our knowledge has not been investigated previously. Most of previous works regarding background/foreground segmentation are based on clustering techniques and have difficulty for the case where the background color is smoothly changing  over a relatively large intensity range, since the background color could be similar to foreground color in some regions. But using the proposed method we can easily segment this kind of images. An example of this case is shown in Figure 1 where we compared the segmentation result by our algorithm with hierarchical k-means clustering used in DjVu \cite{djvu}.

\begin{figure}[1 h]
\begin{center}
\hspace{-0.1cm}
    \includegraphics [scale=0.15] {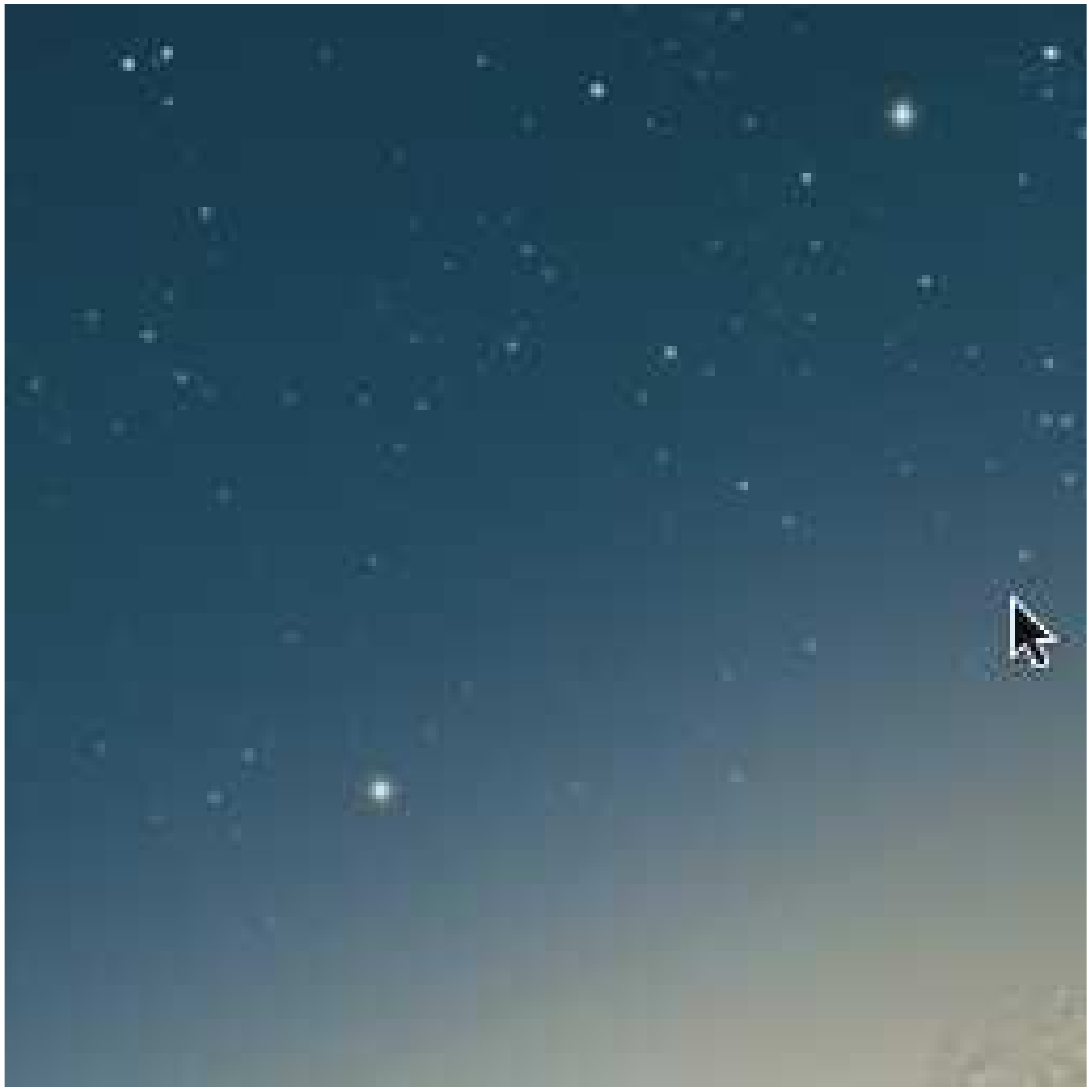}
\hspace{0.2cm}	\includegraphics [scale=0.2] {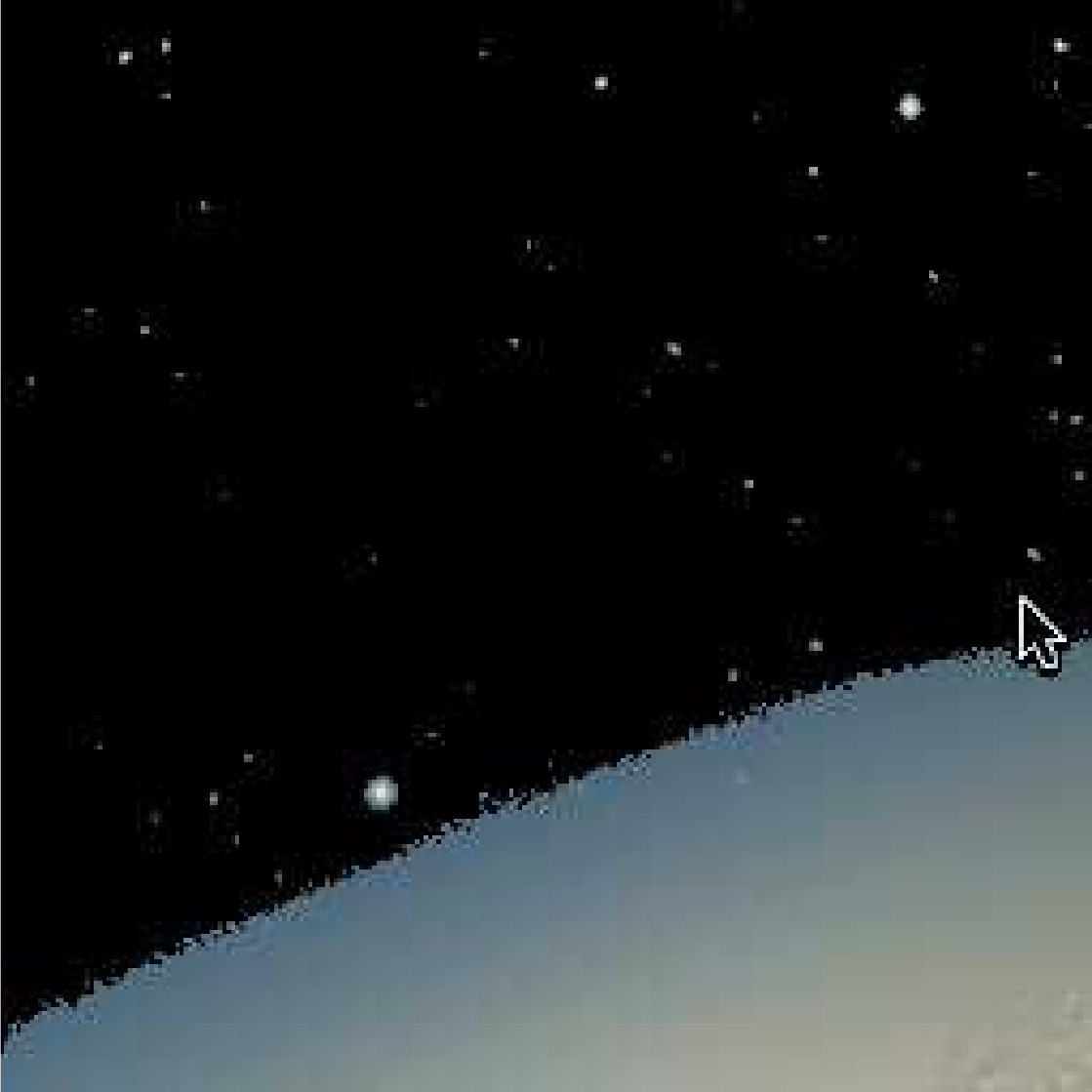}
 \hspace{0.2cm}  \includegraphics [scale=0.15] {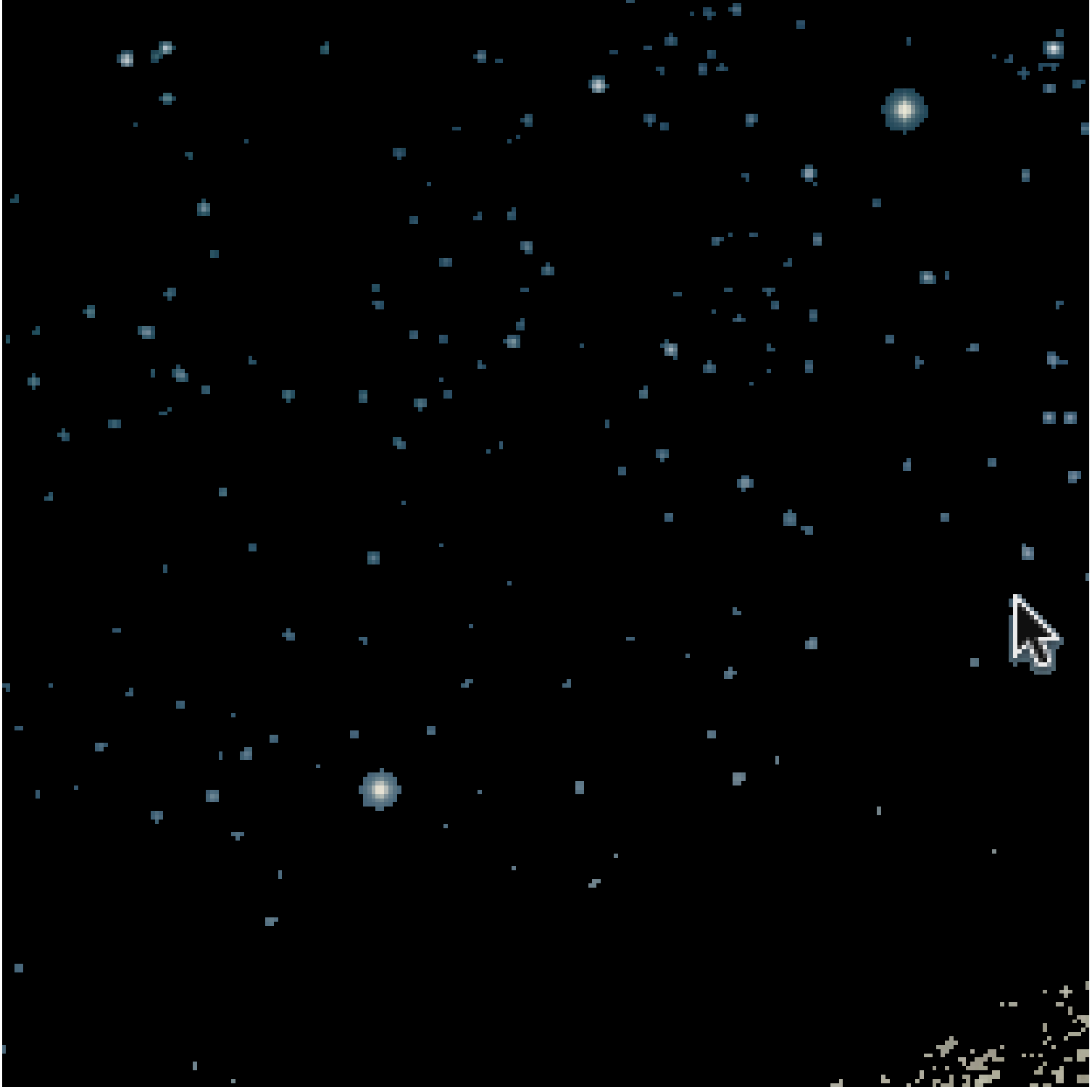} 
\end{center}
  \caption{Segmentation result for a block with smoothly changing background. The left, middle and right images denote the original image, segmented foreground by hierarchical clustering and proposed algorithm respectively.  }
\end{figure}

The previous approaches also have difficulty for the case where some texts in the foreground have similar color to background, our algorithm on the other hand can correctly segment these cases. An example of this case is shown in Figure 2.

\begin{figure}[2 h]
\begin{center}
\hspace{-0.1cm}
    \includegraphics [scale=0.15] {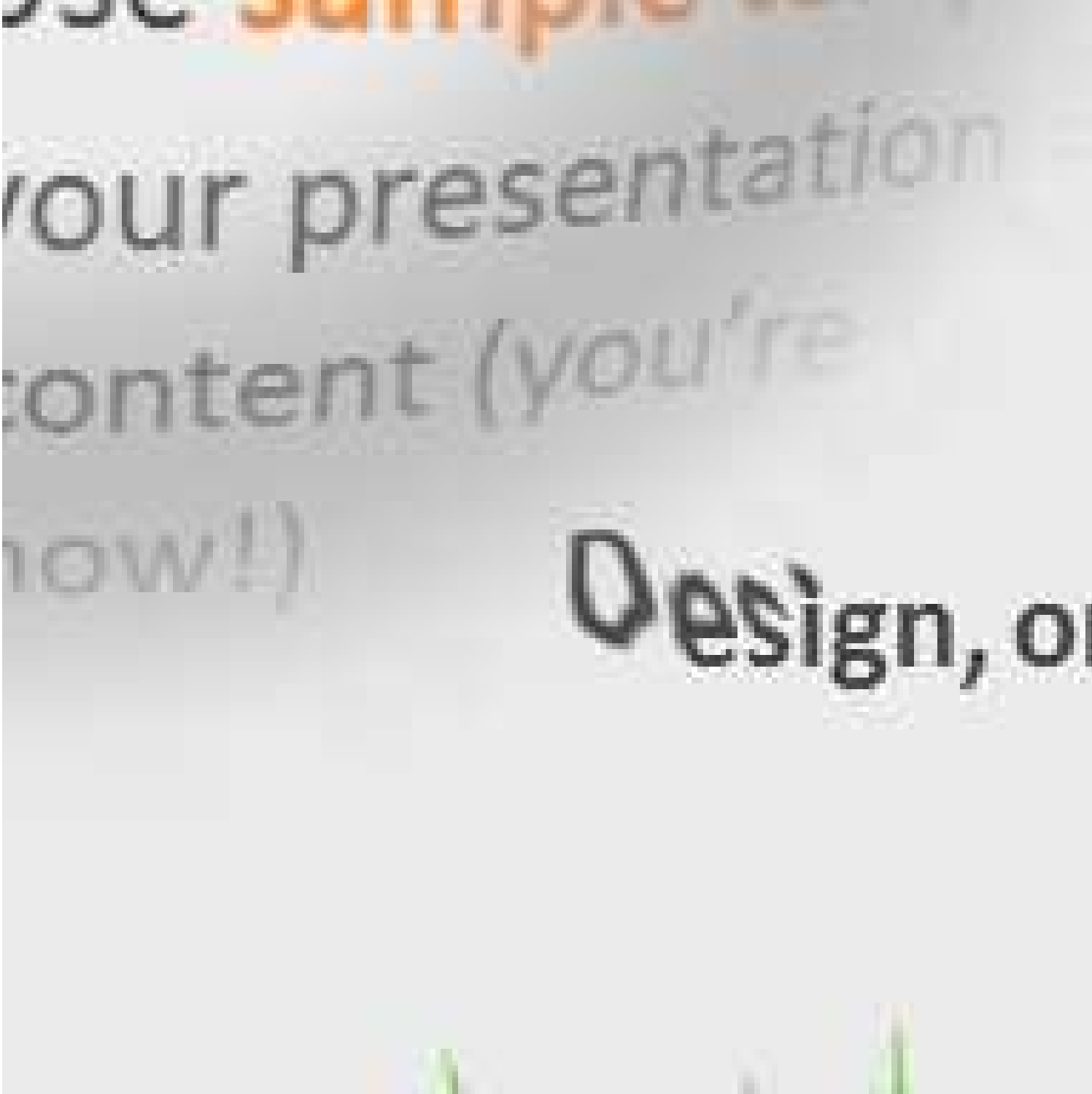}
\hspace{0.2cm}	\includegraphics [scale=0.218] {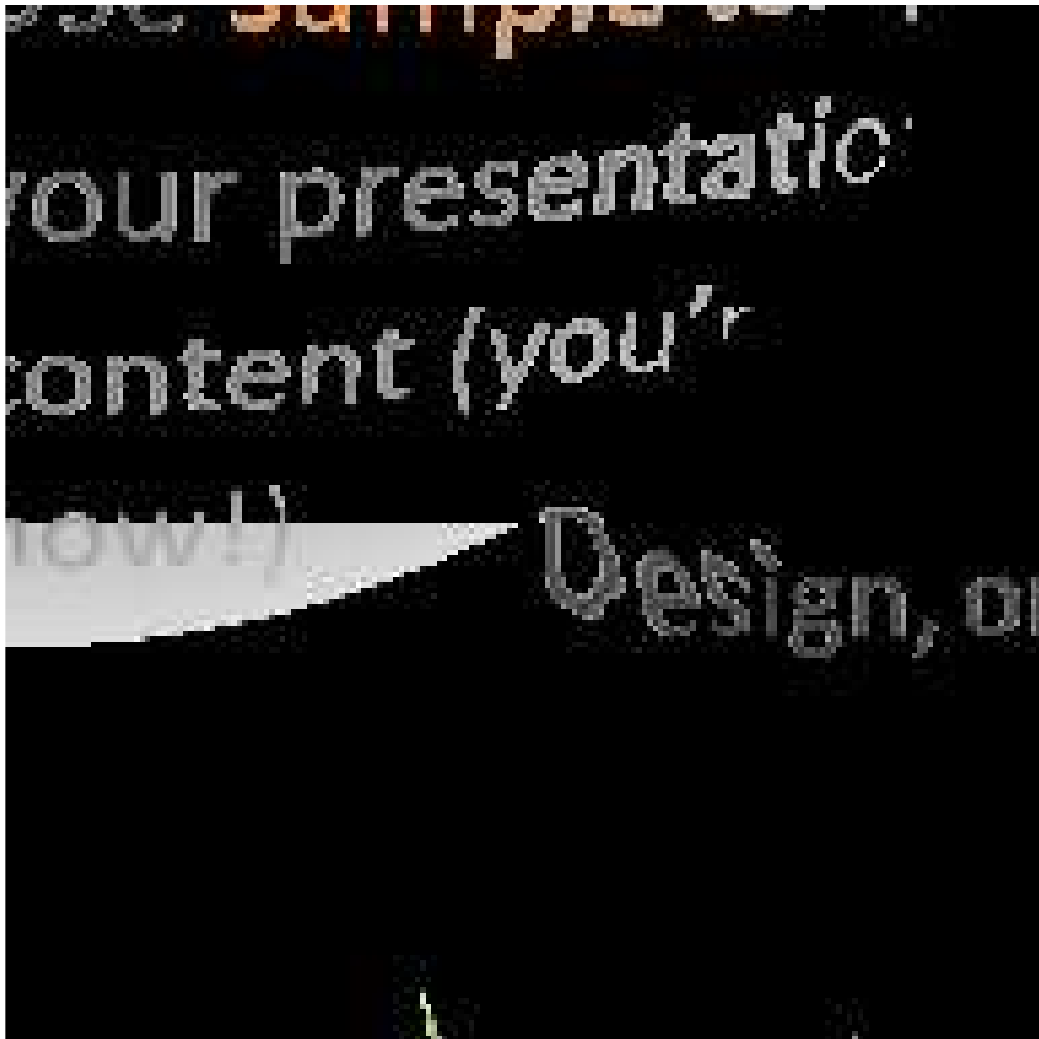}
  \hspace{0.2cm}  \includegraphics [scale=0.15] {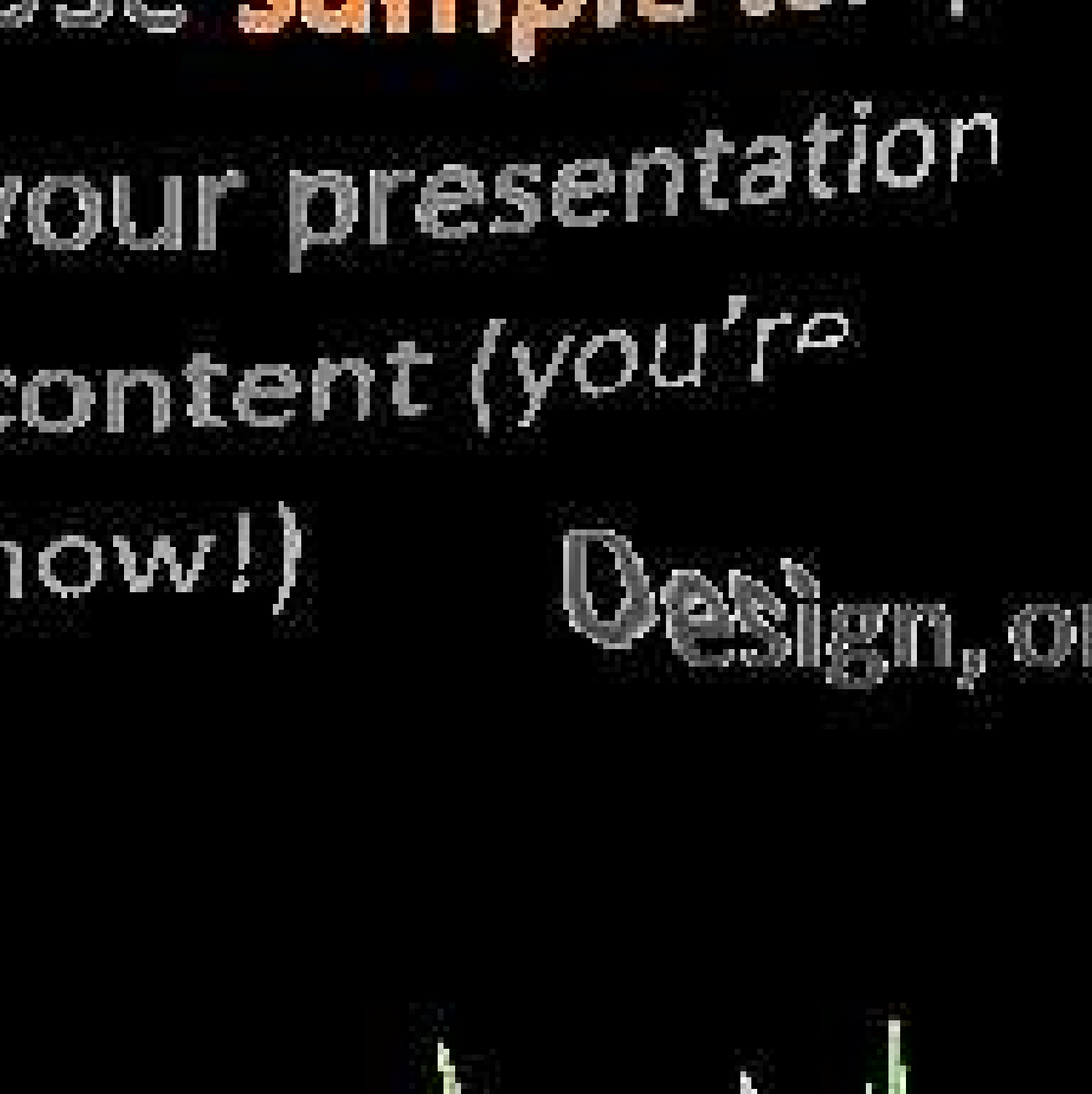} 
\end{center}
  \caption{ Segmentation result for a block with background color similar to foreground shown on the left. The middle and right images are segmented foreground by hierarchical clustering and proposed algorithm respectively. }
\end{figure}

The definition of background and foreground could be different depending on the application. In this paper, background is defined as the smooth part of the image and the foreground are pixels which do not follow the smooth variation pattern of the background. For screen content and mixed document, the foreground typically correspond to text and line graphics. The idea of segmenting an image for better compression was successfully exploited in the DjVu algorithm for scanned document compression \cite{djvu}. The performance of this algorithm highly depends on the segmentation accuracy. 
In DjVu, segmentation is accomplished by applying the k-means clustering algorithm with k=2 on blocks in multi-resolution. It first applies the k-means clustering algorithm on a block with a large size to obtain foreground and background colors and then uses them as the initial foreground and background colors for the smaller blocks in the next stages. Each time, it finds the the foreground (background) color as the weighted average of the foreground (background) color in the current stage and the one in the previous stage. A post-processing step is finally applied to refine the segmentation. Although this segmentation algorithm works pretty well for the images where the foreground and background have a large contrast, it has difficulties for the cases where foreground contains several components with different colors and some of them have similar color to the background.

In another work \cite{spec}, Lin proposed a two-step procedure for background/foreground segmentation. In the first step, for each block of size 16*16 they decide whether it belongs to background or foreground by thresholding the number of colors in each block. If the number of color is more than 32, they classify that block as background. The first stage produce a coarse segmentation, since some of the pixels in background could belong to foreground, then in the second stage they further refine the segmentation by extraction texts and graphics from background. This algorithm suffers from the first stage where any block with the number of colors less than 32 is assigned into text/graphics blocks which is not always true.

In \cite{check}, the authors proposed to use a morphological filtering approach to separate background from foreground in check images. Although the performance is good, it may not be applicable to all types of mixed content images with different structures.

There are also some other segmentation techniques that are based on adaptive thresholding and they do not usually perform well for blocks with several foreground components.

We note that none of the prior approach explicitly make use of the fact that the background is typically smoothly varying, even though it may have a large dynamic range (e.g. a linear change from black to white). In fact any clustering-based segmentation method would fail when the background has a large dynamic range, as it will split these pixels into separate clusters.
Here we proposed a segmentation algorithm based on robust regression techniques to overcome the problems of previous segmentation algorithms. In this approach we try to model the background part of the image with a smooth function by fitting a smooth model to the intensities of the majority of the pixels in each block. And any pixel whose intensity could be predicted well using the derived model would be considered as background and otherwise it would be considered as foreground. The smooth model for intensities is found by a robust regression algorithm, RANSAC (random sample consensus) \cite{RANSAC}.

The structure of this paper is as follows: Section 2 presents the general idea of the proposed segmentation method. The procedure of basis selection for background representation is explained in Section 2.1. Section 2.2 briefly describes robust regression technique and its application for segmentation. Section 2.3 presents the detail of RANSAC algorithm for segmentation. The final segmentation algorithm is discussed in Section 3. Section 4 provides the experiment result for this algorithm on several challenging test images. Section 5 presents some preliminary results regarding the application of proposed scheme for principle line extraction in palmprint images.

\section{The proposed algorithm for foreground/background segmentation}

The previous approaches for foreground/background segmentation are mainly based on color clustering and adaptive thresholding. Here we look at this segmentation problem from regression point of view. We assume that if an image block only consists of background, it should be well represented with a few smooth basis functions. By well representation we mean that the approximated value at a pixel with the smooth functions should have an error less that a desired threshold at every pixel. But if an image block has some foreground pixels overlaid on top of a  smooth background, and these foreground pixels occupy a relatively small percentage of the block, then the fitted smooth function will not represent these foreground pixels well. Based on this notion we try to represent an image block as a linear combination of $K$ smooth functions $\sum_{k=1}^K \alpha_k P_k(x,y)$, where $P_k$ denotes a 2D smooth function. To find the model weights, $\alpha_k$, we should solve a regression problem to fit this model to pixels' intensities. Since the foreground pixels cannot be modeled with this smooth representation they would usually have a large distortion by using this model. Therefore the foreground segmentation task simplifies into finding the set of outlier pixels which cannot be approximated well using this model. Now some question arises here:
\begin{enumerate}
\item What are the right basis functions, $P_k(x,y)$, that can represent the background layer accurately and compactly.
\item How can we solve this regression problem such that the model parameters are not affected by foreground pixels, especially if we have many foreground pixels.
\end{enumerate}
For the first question we can select the right set of basis functions by training over several pure background images. For this part, we performed an experiment by collecting several background images and trying different basis functions and it turns out using DCT basis produces better results than others. The details of this experiment is explained in Section 2.1. 
There are also some approaches designed to extract suitable basis functions in the presence of outliers \cite{jons1}.

For the second question we proposed a robust regression algorithm which can find the right model for a set of data in presence of outliers. This approach is explained in Section 2.2. The proposed approach has some connections with subspace recovery problem \cite{rahmani1}, \cite{jons2}, \cite{rahmani2}.
We can also use a sparse decomposition algorithm to find the outliers. This sparse decomposition is briefly explained in the Appendix and more detailed study is left for future studies.

\subsection{Basis Selection}
As explained earlier, basis are used to represent the background layer of the image, therefore they should be chosen such that they are suitable for background representation. There are two ways to select a set of basis which best represents a signal or an image. One is to collect a sufficient amount of training data and then use Karhunen-Loeve transform to find the best set of basis \cite{KLT}. The other way is to use a set of pre-designed basis which are suitable for background representation. 
Here we tried two set of basis functions which are known to be very efficient for smooth images, the DCT basis and the orthonormal polynomials. The two dimensional DCT basis are products of 1D DCT basis, and are well known. To derive 2D orthonormal polynomials over an image block of size NxN, we can use the simple polynomials $f_n(x)=x^n$ and orthonormalize them using Gram-Schmidt process to get N orthonormal basis by calculating the values of $f_n(x)$ at $x= 1,2,...,N$.
After deriving the 1D orthonormal polynomials, we can reconstruct 2D orthonormal polynomials as the outer-product of 1D bases.

To compare DCT and orthonormal polynomials, we collected several smooth background images, extracted all blocks of size 64*64 and tried to represent those blocks with the first K polynomials and DCT basis functions in zigzag order as $\sum_{k=1}^K \alpha_k P_k(x,y)$. If we look at vectorized version of this equation we will have $F=P \alpha$ where $P$ is a $N^2 \times K$ matrix with the k-th column corresponding to $P_k(x,y)$ for all $(x,y)$ in the block. Then $\alpha$ can be found by minimizing the mean square error as: $\alpha= (P^T P)^{-1} P^T F$. Then we tried to use this model to predict pixels' intensities and find the MSE for each block,. The reconstruction RMSEs (root MSE) as a function of number of used bases, $K$, for both DCT and polynomials are shown in Figure 3. As we can see DCT has slightly smaller RMSE, so it is preferred over orthonormal polynomials.

\begin{figure}[3 h]
\begin{center}
    \includegraphics [scale=0.35] {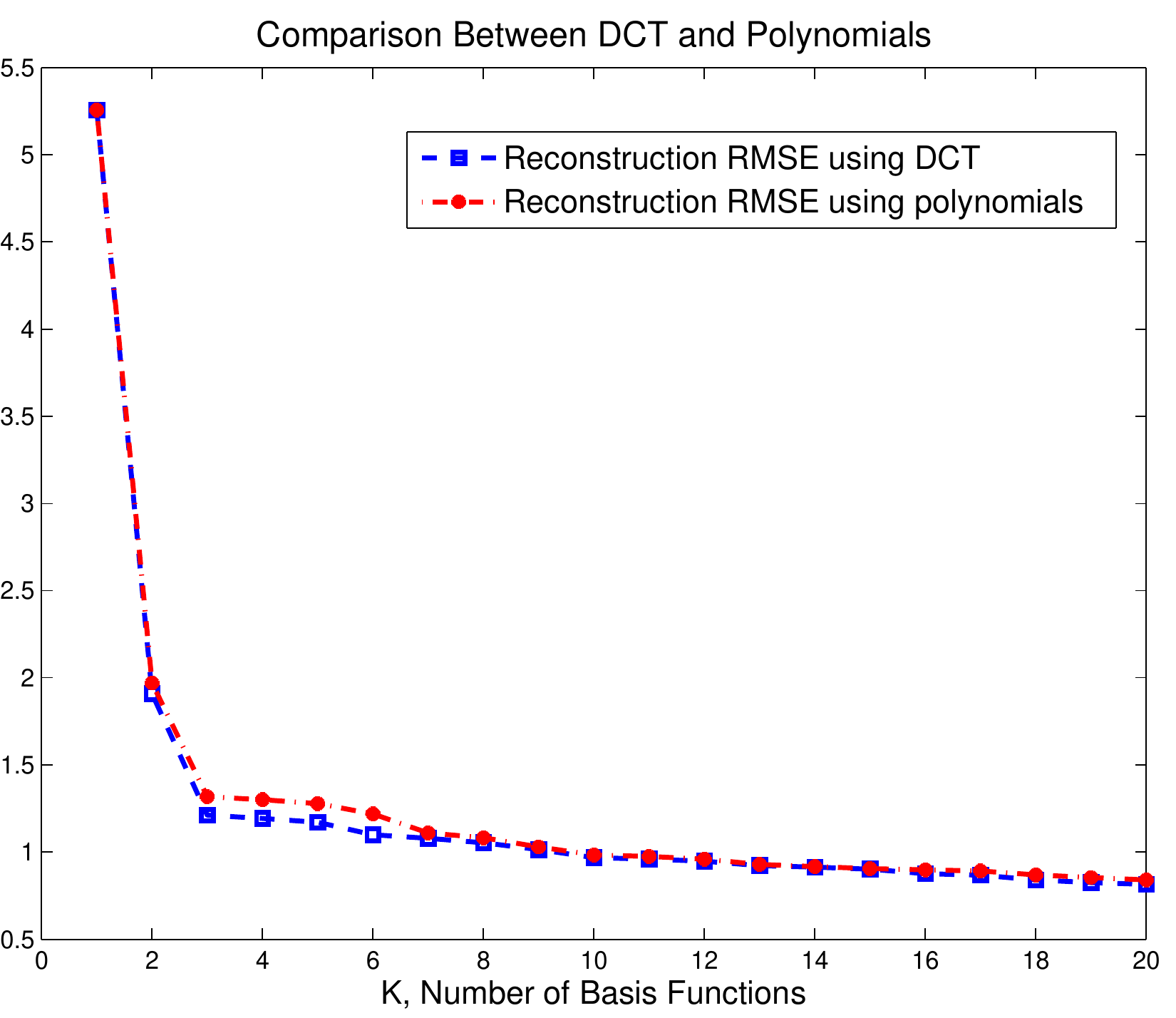}
\end{center}
  \caption{Background reconstruction MSE as fucntion of number of bases K }
\end{figure}

The number of DCT basis functions, $K$, is chosen as the minimum number such that the reconstructed background has a PSNR about 45 dB which results in a good reconstructed image quality (here $K=10$). In general, the choice and number of basis functions depend on the application and characteristics of the typical background  patterns in the application.

\subsection{Robust Regression and RANSAC}
Robust regression is a form of regression analysis which is developed to overcome some limitations of traditional algorithms \cite{robustregression}. The performance of most of the traditional regression algorithms can be significantly affected if the assumptions about underlying data-generation process are violated. Most of the traditional regression algorithms are highly sensitive to the presence of outliers. The outlier can be thought as any data-point or observation which does not follow the same pattern as the rest of observations. The robust regression algorithms are designed to find the right model for a dataset even in the presence of outliers. These robust estimators try to fit a model only to the inliers. They basically try to remove the outliers from dataset and use the inliers for model prediction.

There are two general approaches to solve robust regression problem:
\begin{enumerate}
\item Estimating the model's parameters using a robust cost function and then detect the outliers as the data points which deviate from the right model by a large value.
\item First indentify the outliers, remove them from dataset and then use a usual regression algorithm to find the model parameters.
\end{enumerate}
M-estimators \cite{M_es} and RANSAC \cite{RANSAC} are two popular robust regression algorithms. M-estimators are the generalized version of maximum likelihood estimator, they replace the log-likelihood function by a robust cost function. RANSAC is an iterative approach which performs the parameter estimation by minimizing the number of outliers. The outliers are defined as the points which have a prediction error more than a pre-defined threshold. RANSAC and its modified versions such as MSAC and MLESAC are perhaps the most popular robust estimators in computer vision \cite{RR_vision}. In this work, we used RANSAC as a robust estimator to perform background/foreground separation, but other robust regression algorithms can also be used for this task. RANSAC repeats the following algorithm for robust regression:
\begin{enumerate}
\item Select a random subset of the data with the minimum size which is required to determine all the parameters of the model. Let us call this subset as the minimal sample set (MSS).
\item Fit a model to the points in the minimal sample set.
\item Test all other data points against the fitted model. Those points that can be predicted sufficiently well according to some cost function, will be considered as the inliers (consensus set).
\item The model will be considered reasonably good if the number of inliers is sufficiently large.
\end{enumerate}
We have to repeat this procedure a fixed number of times, each time producing a model with a corresponding consensus set. Finally we use the model that has the largest consensus set. After this procedure is finished, the model could be improved by re-estimating it using all inliers. 

\subsection{RANSAC for Background/Foreground Segmentation}
As stated earlier, our foreground segmentation algorithm can be simplified into a problem of fitting the pixel intensities in each block into a smooth function in the presence of outliers. Those pixels which can be well represented using the fitted smooth model will be considered as inliers or background and those that cannot be well represented using this smooth model will be regarded as outliers or equivalently foreground. 
Suppose we have a N*N block of image $F: \{1,2,3,...,N\} \times \{1,2,3,...,N\} \rightarrow G$ where $G$ denotes the set of all grayscale levels (intensities), here $G=\{0,1,2,...,255\}$ and we want to fit a set of $K$ low frequency DCT functions to that block. The 2-dimensional DCT function is defined as:
\begin{equation}
P_{u,v}(x,y)= \beta_u \beta_v cos((2x+1)\pi u/2N) cos((2y+1)\pi u/2N) 
\end{equation}
where $u$ and $v$ denote the frequency of the basis and $x$ and $y$ denote spatial coordinate of the image pixel. Then the goal is to find $\alpha_k$'s for each block such that for any pixel at $(x,y)$ belonging to background, its intensity can be approximated as:
\begin{equation}
F(x,y)= \sum_{k=1}^K \alpha_k P_k(x,y)
\end{equation}
As we can see, this is very similar to linear regression problem, where we want to predict the intensity of $N^2$ pixels, $F(x,y)$, as a linear combination of $K$ basis functions. The only difference is that here we want the model not to be affected by outliers (foreground pixels). 

RANSAC can be used to solve this problem. The proposed RANSAC algorithm for foreground/background segmentation of a block of size N*N is as follows:

\begin{enumerate}
\item Select a subset of K randomly chosen pixels. Let us denote this subset by $S$.
\item Fit the model $\sum_{k=1}^K \alpha_k P_k $ to the pixels $F(x,y) \in S$ and find the $\alpha_k$'s. Since the number of pixels equals to the number parameters, $\alpha_k$'s, we can find a unique solution here.
\item Test all $N^2$ pixels $F(x,y)$ in the block against the fitted model. Those pixels that can be predicted with an error less than $\epsilon$ will be considered as the inliers (consensus set).
\item Save the consensus set of the current iteration if it has a larger size than the previous one.
\item Repeat this procedure M times.
\end{enumerate}
After this procedure is finished, the pixels in the consensus set will be considered as inliers or equivalently background. 
The final result of RANSAC can be refined by refitting over all inliers once more. There are some parameters which needed to be tuned. 
Here we have chosen $N=64$ which is the largest coding unit size in HEVC standard \cite{HEVC}. For this value of $N$ the number of required basis for background is chosen to be $K=10$ as explained in Section.2.1. The maximum allowed inlier distortion is adaptive and it varies with range of pixel intensities within each block, specifically $\epsilon=1+0.22 R$ where $R$ denotes the range. The maximum number of iteration is chosen to be 1000. This is determined so that the probability of not finding the right model is $\frac{1}{10^{12}}$ using RANSAC algorithm \cite{RANSAC} (assuming average inlier ratio to be 0.7 and K=10).

The segmentation results by RANSAC is usually very good, but it is computationally demanding. For blocks that can be easily segmented with other methods, RANSAC may be an overkill. Therefore, we propose a segmentation algorithm which has different modes in the following section.

\section{The Proposed Segmentation Algorithm}
We propose a segmentation algorithm which mainly depends on RANSAC but it has some initial modes to first check if a block can be segmented using some simpler approaches and it goes to RANSAC only if the block cannot be segmented using those approaches. These simple cases belong to one of these groups: completely flat block, smoothly varying background only and text/graphic overlaid on constant background.
 
Completely flat blocks are those in which all pixels have the same value. These kind of blocks are common in screen content images. They could be either a constant background or a constant region inside a text/graphics. Therefore they can be declared as background or foreground based on their neighboring blocks' background color. We basically compare their colors with all their neighboring background colors. If we could find at least one neighbor with a background color close enough to the current block's color (difference less than 10 out of a range of 0 to 255), it would be segmented as background.

Smoothly varying background only is a block in which the intensity variation over all pixels can be modeled well by a smooth function. Therefore we try to fit $K$ DCT basis to all pixels using least square fitting. If all pixels of that block can be represented with an error less than a predefined threshold (here 3, for pixel intensity $\in \{0,1,...,255\}$ ), we declare it as smooth background.

The last group of simple cases is text/graphic overlaid on constant background. An example of this kind of images is shown in Figure 4. The images of this category usually have zero variances (or very small variances) inside each connected component. For example, all letters have zero variance in the image in Figure 4. These images usually have a limited number of different colors in each block(usually less than 10) and the intensities in different parts are very different. Therefore they can be easily detected. The segmentation method for these images is based on percentage of colors and neighboring blocks information. First we calculate the percentage of each different color in that block and the one with the highest percentage may be chosen as background and the other ones as foreground. The background color needed to be confirmed by neighboring blocks' background color. If it is not consistent, we should check this procedure for the color with second highest percentage and so on.

\begin{figure}[4 h]
\begin{center}
    \includegraphics [scale=0.45] {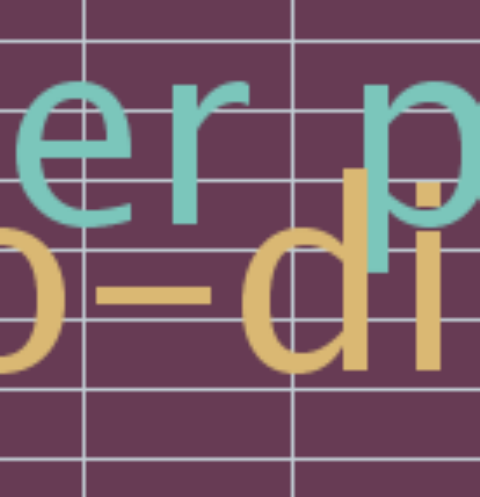}
\end{center}
  \caption{Text and graphics over constant background}
\end{figure}

The overall segmentation algorithm is summarized as follows
\begin{center}
\begin{tabular}{|p{0.95\linewidth}|}\hline 
\rule{0pt}{5ex}
\vspace{-0.8cm}
\begin{enumerate}
\item If all pixels in the block have the same color intensity (i.e. it is completely flat block), declare the entire block as background or foreground based on the similarity of its color with background and foreground of its neighboring blocks. If not, go to the next step;
\item Perform least square fitting using all pixels. If all pixels can be predicted with an error less than $\epsilon_2$, declare the entire block as background. If not, go to the next step;
\item If the number of different colors is less than a threshold and the intensity range is above 50, declare it text/graphics overlaid background and found the background as explained above. If not, go to the next step;
\item Use RANSAC to segment background and foreground. If the percentage of inliers is more than a threshold $\epsilon_3$ or $N=8$ then all inlier pixels are selected as background. If notgo to the next step;
\item Decompose current block of size $N \times N$ into 4 smaller blocks of size $\frac{N}{2} \times \frac{N}{2}$ and run the segmentation algorithm for all of them. Repeat until $N=8$.
\end{enumerate}
\\\hline
\end{tabular}
\end{center}
An example of advantage of quad-tree decomposition is shown in Figure 5 where the foreground map (a map which denotes the location of foreground pixels) without and with quad-tree decomposition are shown in the middle and right images respectively. For the purpose of illustration $\epsilon_3$ here is chosen such that the difference is more clear. As we can see, we can get much better result compare to the case with no decomposition.

\begin{figure}[4 h]
\begin{center}
\hspace{-0.1cm}
    \includegraphics [scale=0.15] {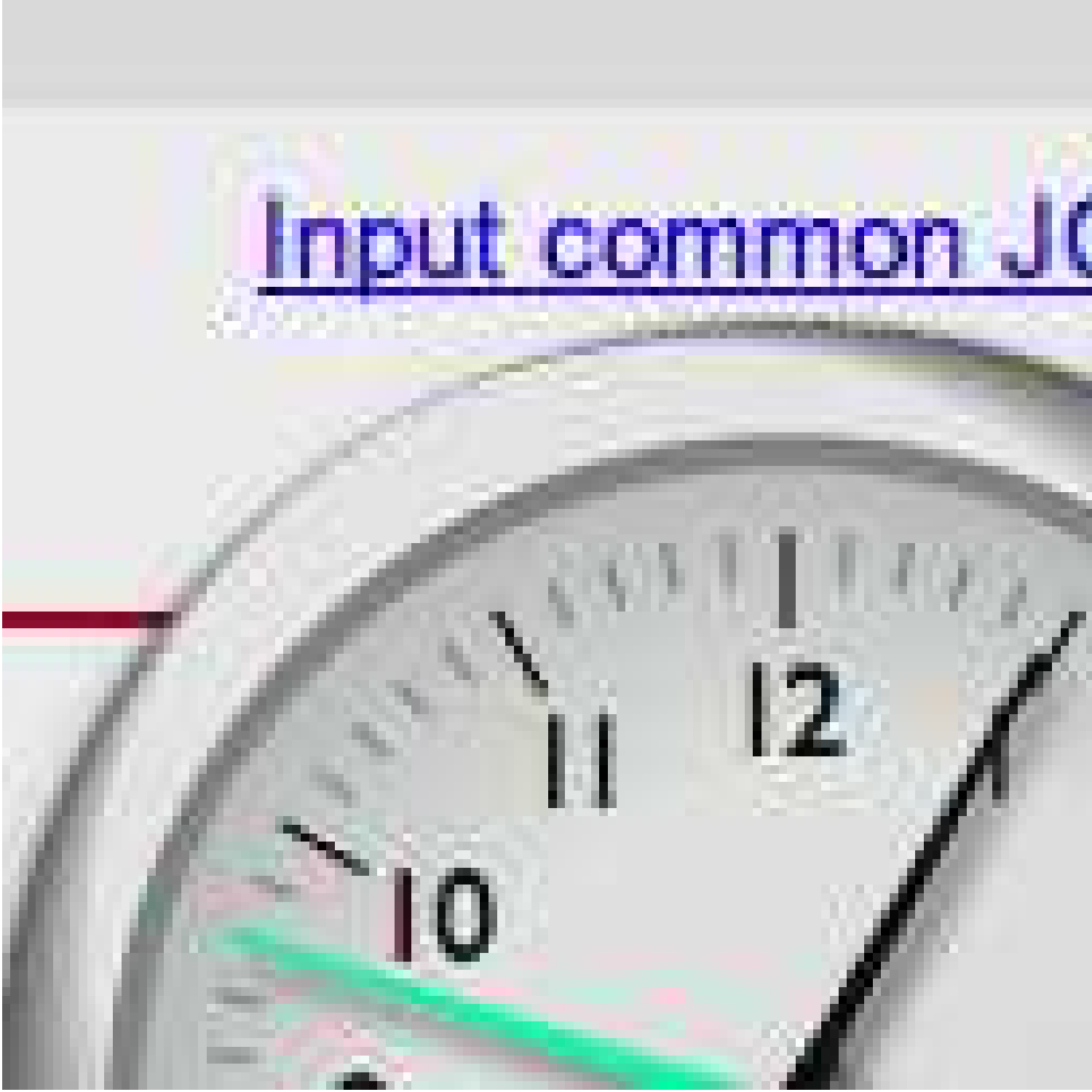}
\hspace{0.2cm}	\includegraphics [width=2.3cm, height=2.3cm] {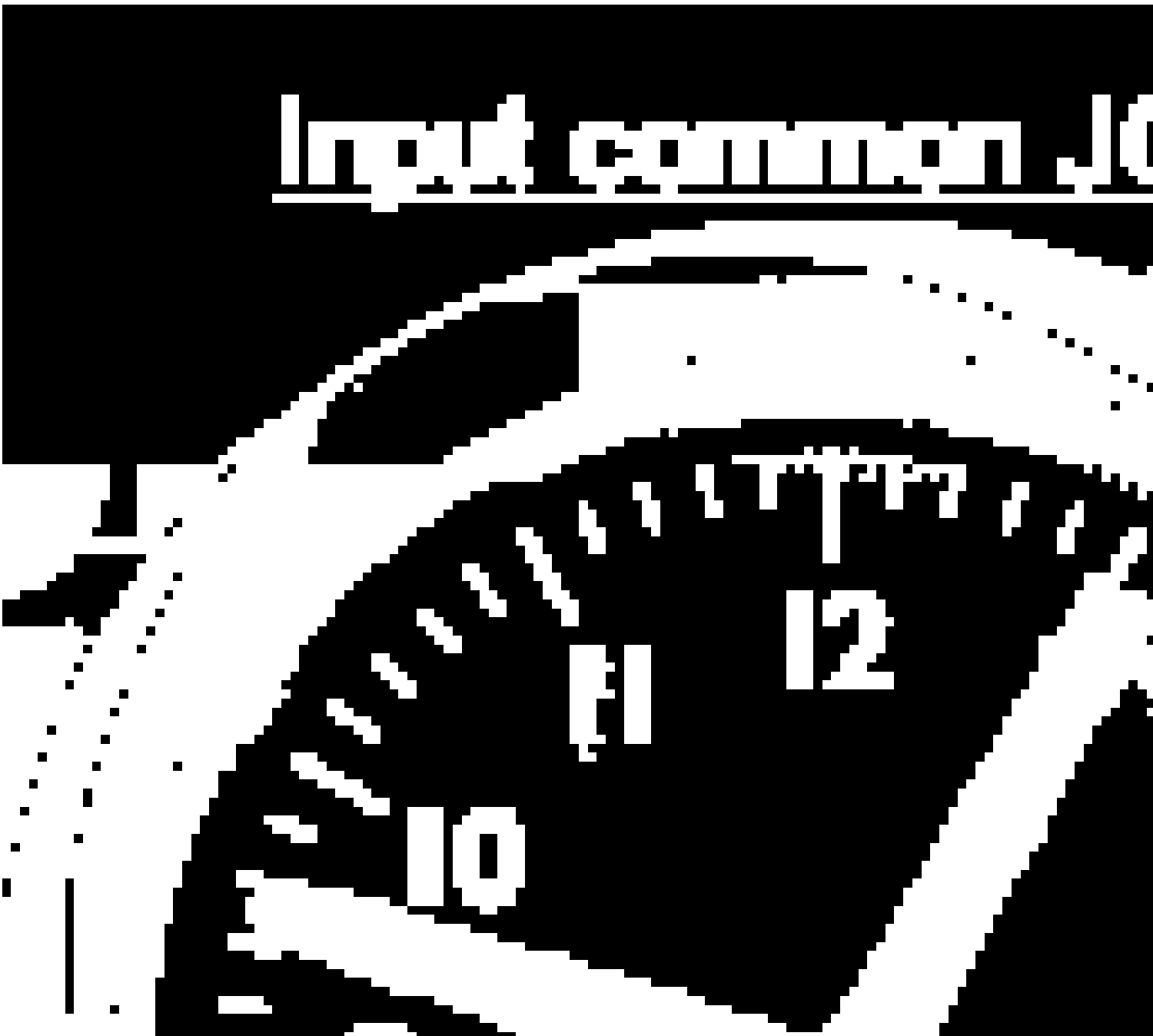}
  \hspace{0.2cm}  \includegraphics [scale=0.15] {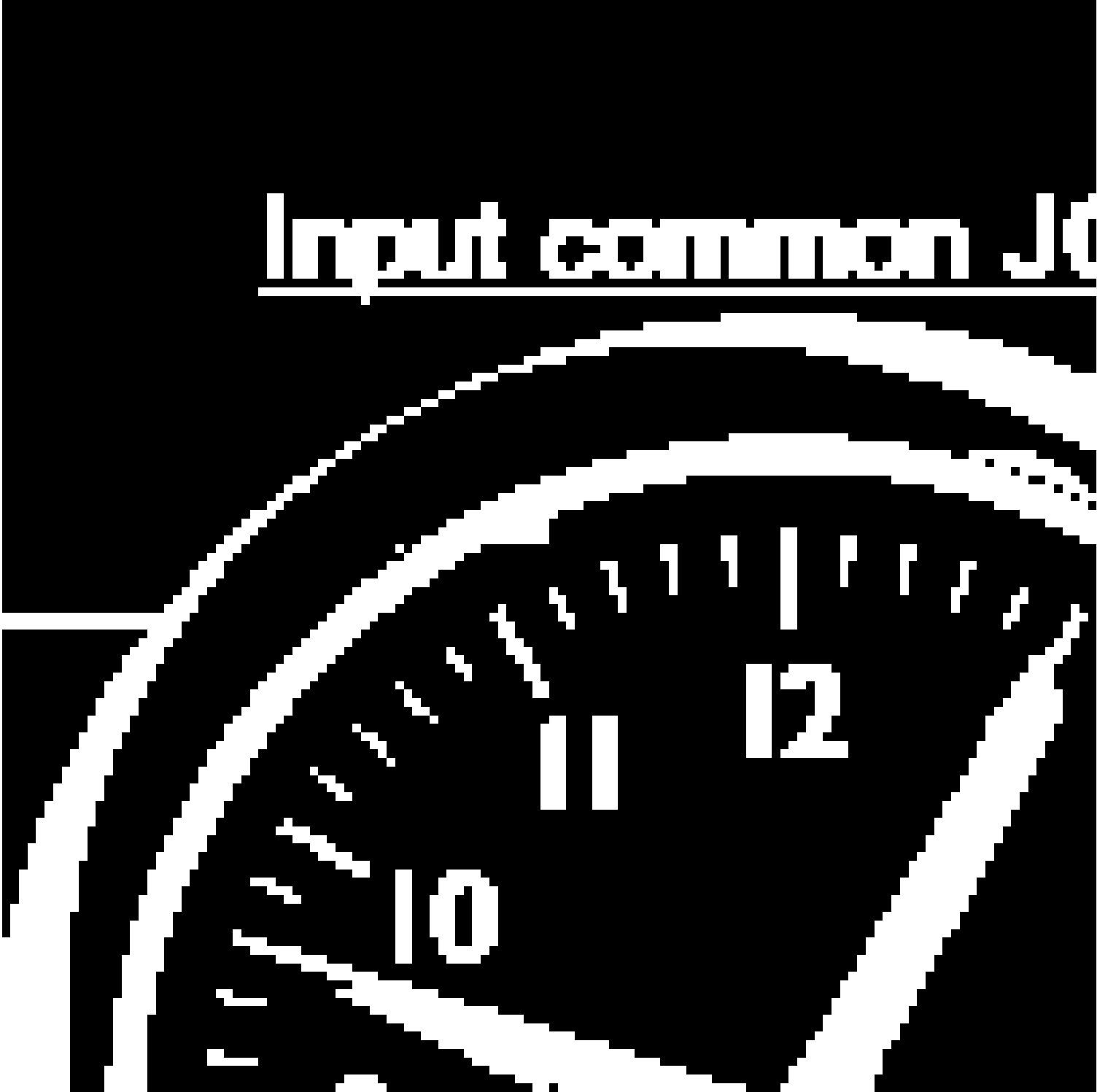} 
\end{center}
  \caption{Segmentation result for a sample image, middle and right images denote foreground map without and with quad-tree decomposition  }
\end{figure}

\section{Results for Foreground/Background Segmentation}
\label{SectionIV}
We have tested our algorithm on several test images form HEVC standard test sequences for screen content coding. They usually consist of text or a line graphics overlaid over smooth background.
In our experiment the quad-tree decomposition can be repeated up to the blocks of size 8*8, $\epsilon_2 =3$, $\epsilon_3 =0.5$
We have compared our algorithm result with hierarchical k-means clustering which is very similar to the segmentation algorithm in DjVu algorithm \cite{djvu}.

For screen content test images, the original image and the foreground maps with two segmentation algorithms are shown. Foreground map is a map in which each element is 1 if its corresponding pixel in the original block belongs to the foreground and zero otherwise.

The first test image is chosen from some part of Slideshow sequence and is shown in Figure 6. The background of this image has a shaded sector which makes it difficult to be segmented. As it can be seen, hierarchical clustering has difficulties with the shaded region inside background and has segmented some parts of background as foreground. It also has problem with the text "you don'" in the upper part of the image, whereas our algorithm segmented all of them correctly.

The second test image is chosen from FlyingGraphics test sequence and it contains texts with different colors where some of them have a color similar to background. The foreground map by hierarchical clustering in DjVu and proposed algorithm are shown in Figure 7. As it can be seen, the hierarchical clustering method cannot detect the small text in the lower part of the image, because it has a color similar to background.

The third test image is chosen from Programming sequence and it has a difficult graphic embedded over smoothly varying background. The segmentation result is shown in Figure 8. It can be seen some of the lines are missing in the segmentation result of hierarchical clustering, whereas the proposed algorithm has detected most of them correctly.

The next test image is taken from MissionControl sequence, the result is shown in Figure 9. Again, the proposed method was able to identify those foreground pixels with similar intensity as the background, which were missed by the clustering approach.

\begin{figure*}[ht]
        \centering
        \vspace{-0.5cm}
        \begin{subfigure}[b]{0.3\textwidth}
                \includegraphics[width=\textwidth]{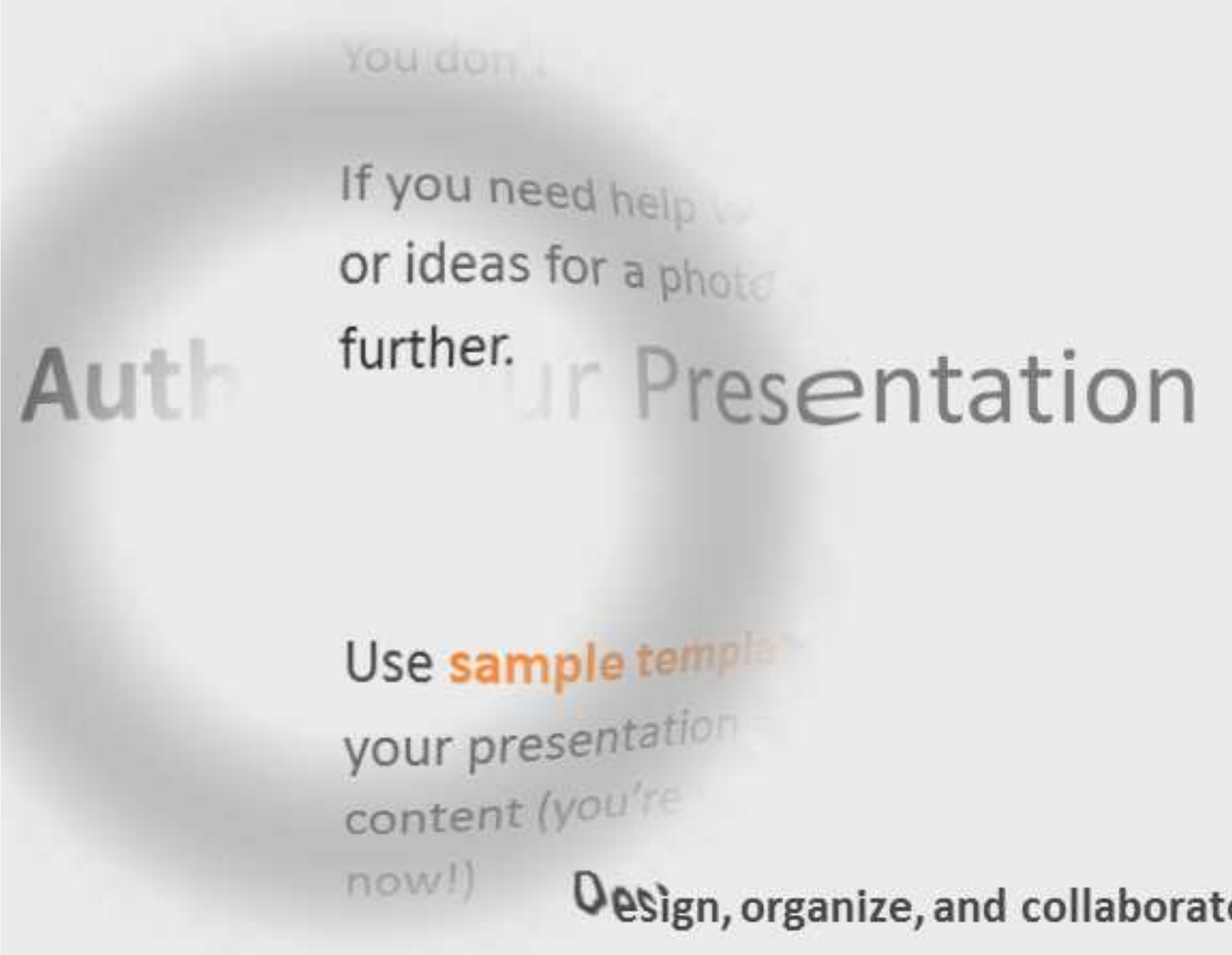}
                                \vspace{-0.5cm}
          \hspace{-1.5cm}    
        \end{subfigure}%
        ~ 
        \begin{subfigure}[b]{0.3\textwidth}
                \includegraphics[width=\textwidth]{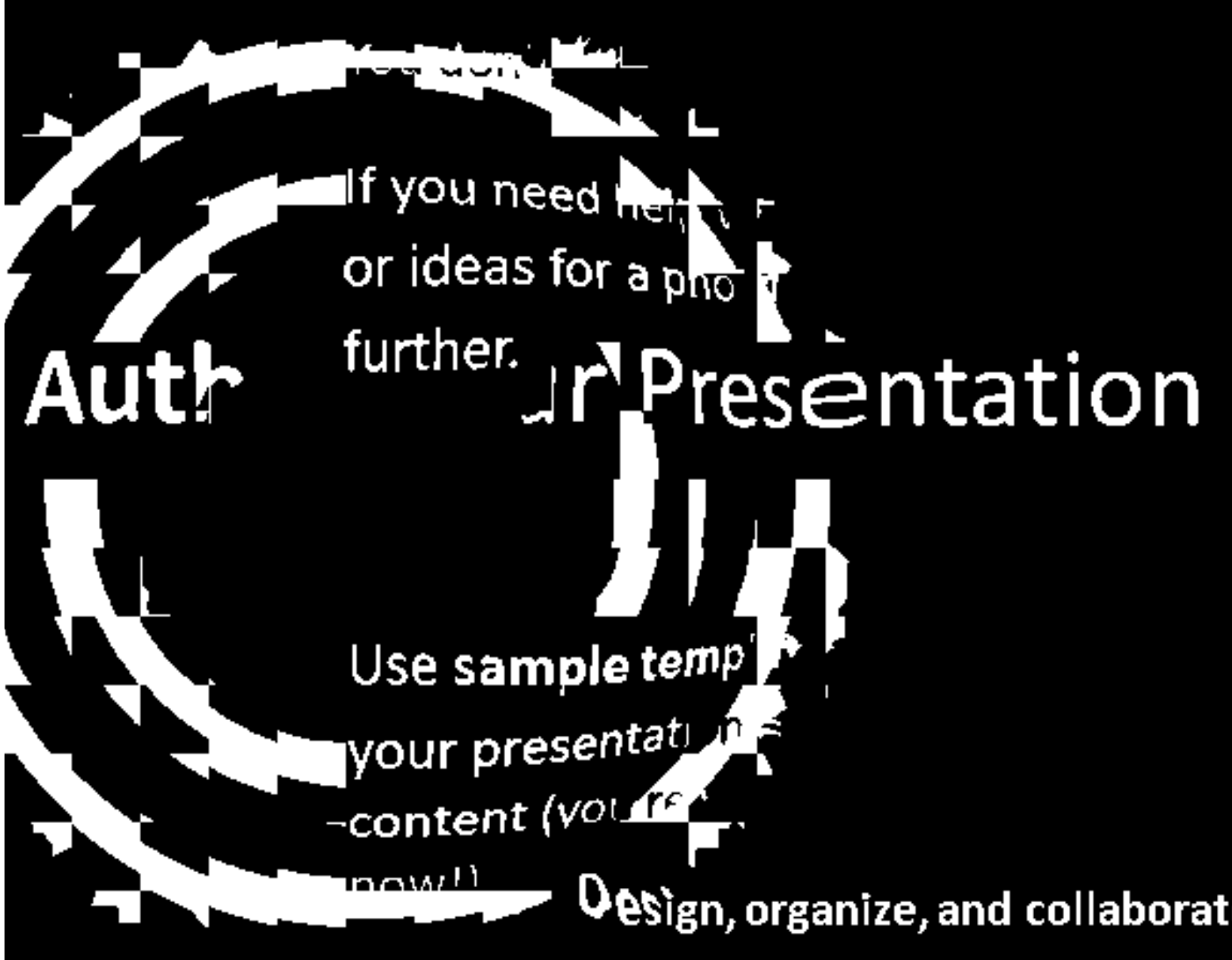}
                \vspace{-0.5cm}
            \hspace{-3cm} 
        \end{subfigure}%
        ~ 
        \begin{subfigure}[b]{0.3\textwidth}
                \includegraphics[width=\textwidth]{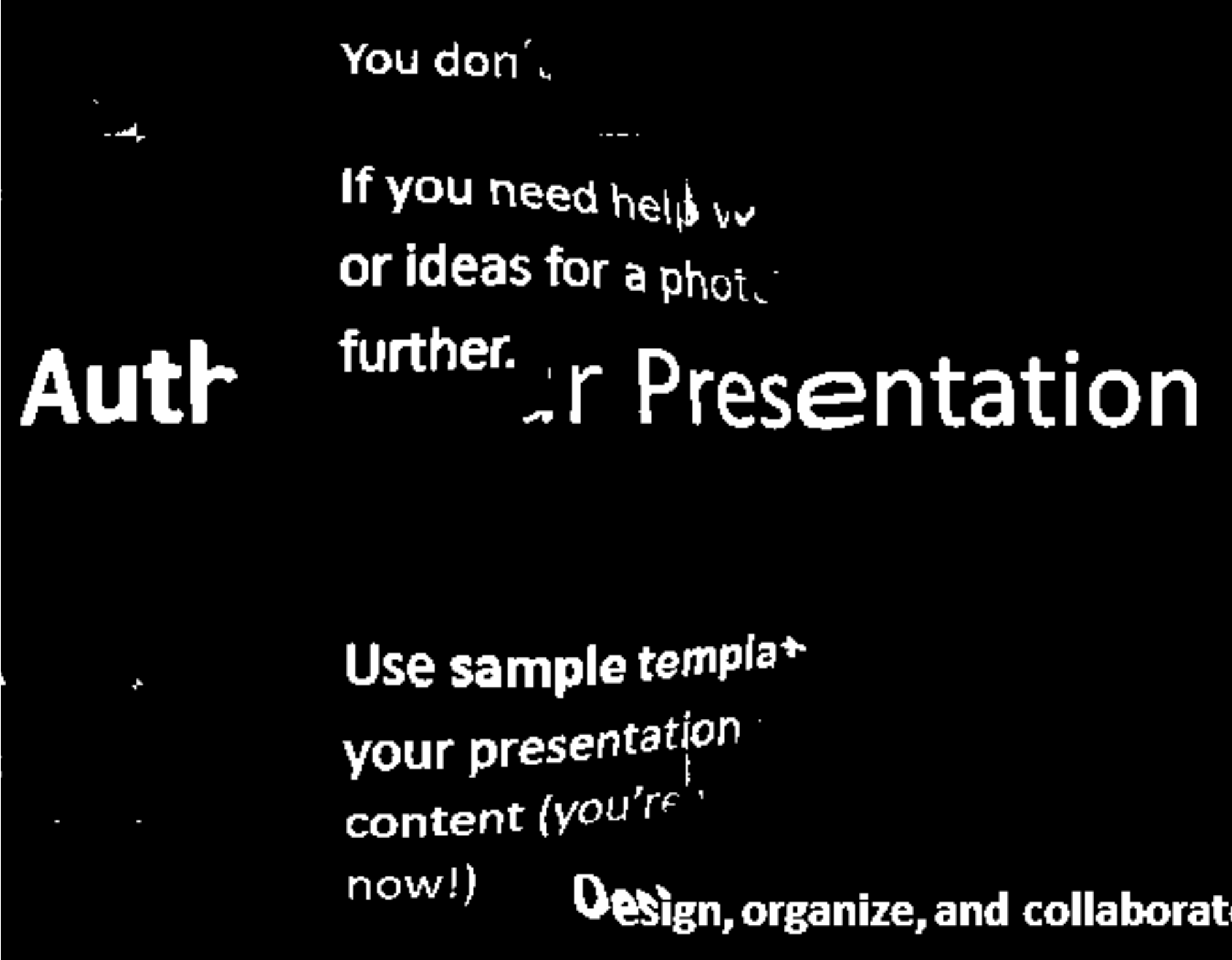}
                 \vspace{-0.5cm}
              \hspace{-4.5cm}
        \end{subfigure}
        \vspace{0.1cm}
        \caption{Segmentation result for Slideshow sequence, the left, middle and right images denote the original, foreground map by hierarchical clustering in DjVu and foreground map by the proposed method respectively}
\end{figure*}

\begin{figure*}
        \centering
        \vspace{-1.1cm}
        \begin{subfigure}[b]{0.33\textwidth}
                \includegraphics[scale=0.32]{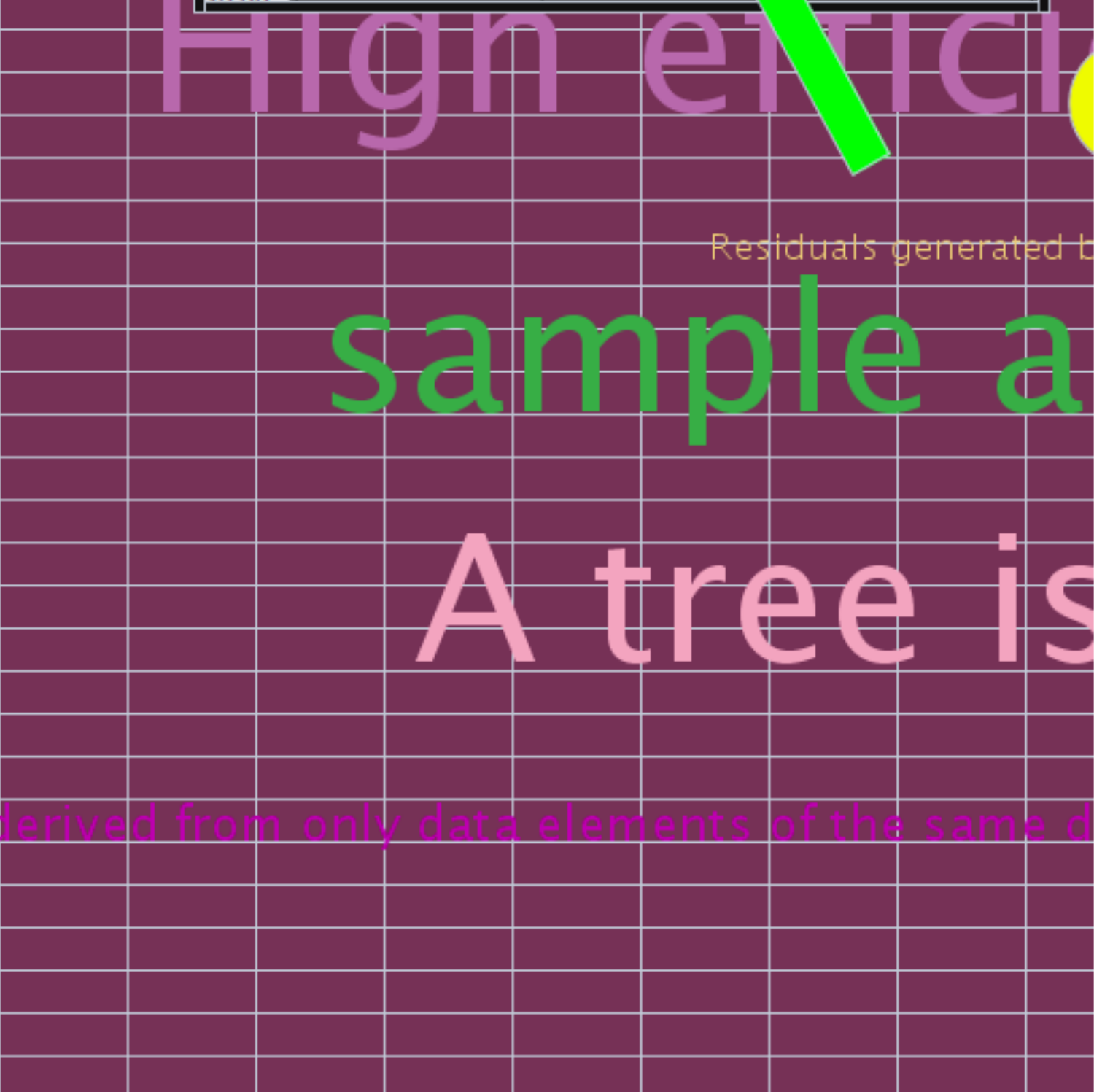}
                                \vspace{-0.5cm}
          \hspace{-1cm}    
        \end{subfigure}%
        ~ 
        \begin{subfigure}[b]{0.33\textwidth}
                \includegraphics[scale=0.373]{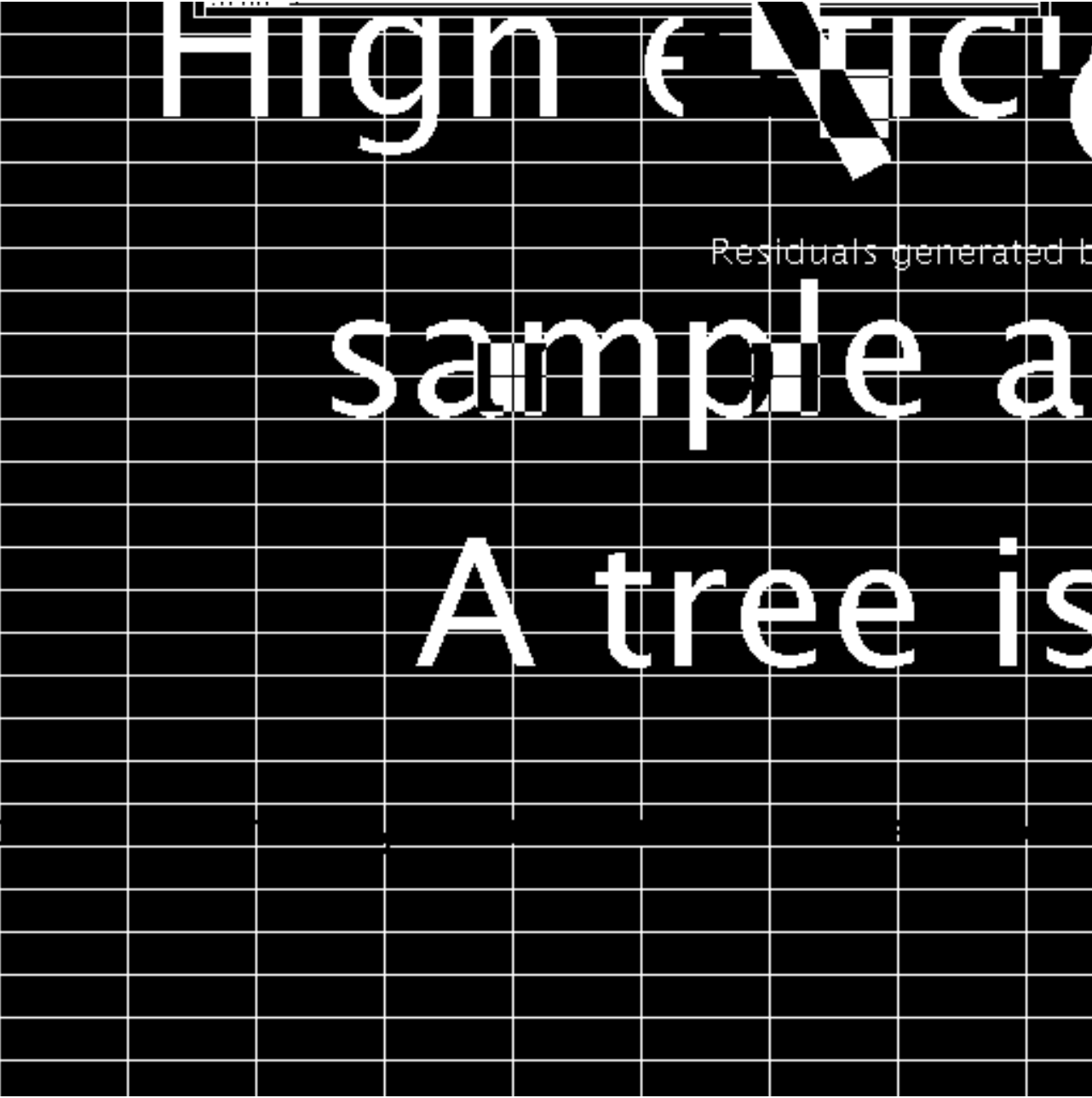}
                \vspace{-0.5cm}
            \hspace{-5cm} 
        \end{subfigure}%
        ~ 
        \begin{subfigure}[b]{0.33\textwidth}
                \includegraphics[scale=0.32]{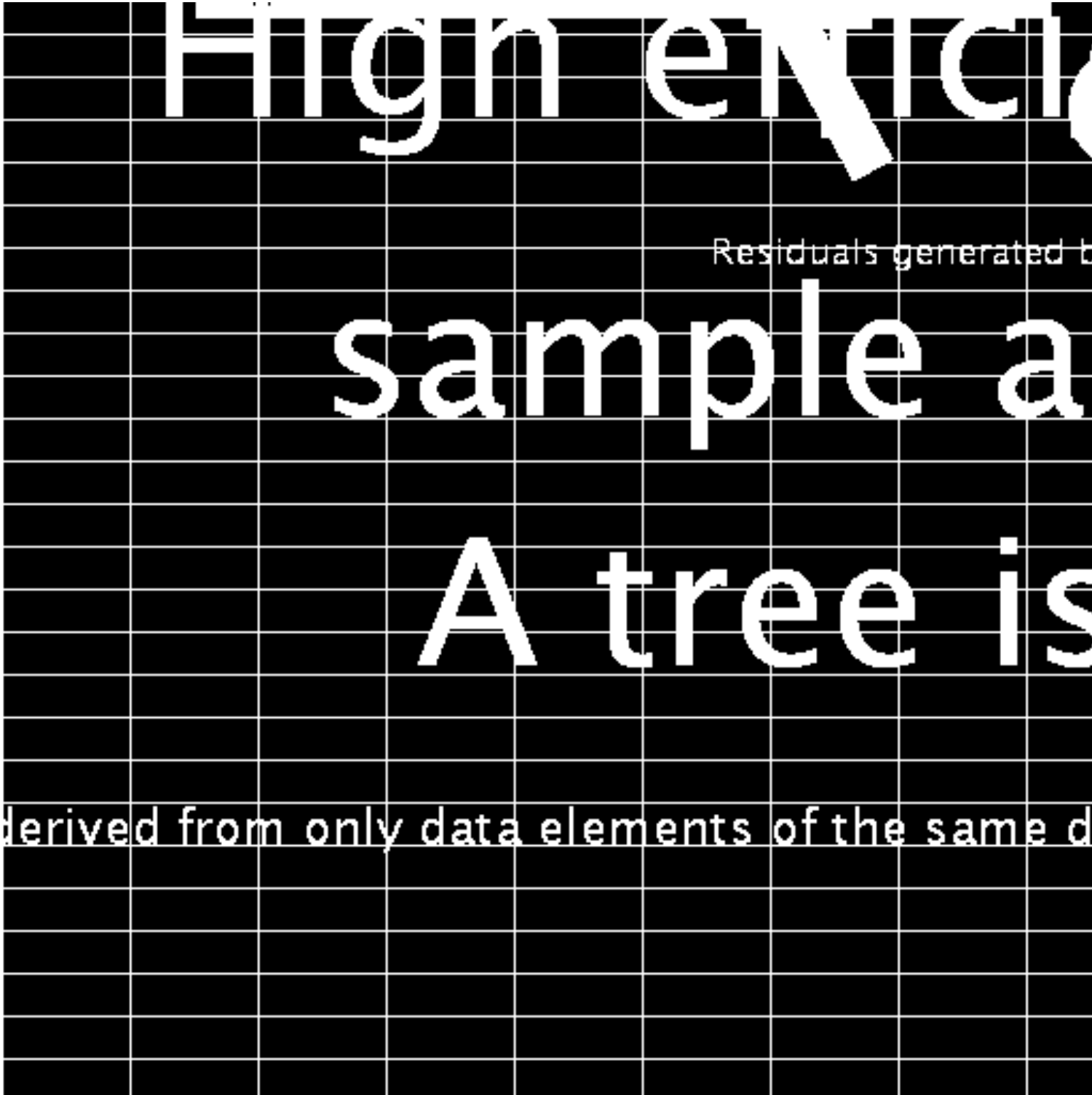}
                 \vspace{-0.4cm}
              \hspace{-7cm}
        \end{subfigure}
        \vspace{0.2cm}
        \caption{Result for FlyingGraphic image, the left, middle and right images denote the original, foreground map by hierarchical clustering in DjVu and the proposed method respectively }
\end{figure*}

\begin{figure*}
        \centering
        \vspace{-1.1cm}
        \begin{subfigure}[b]{0.31\textwidth}
                \includegraphics[width=\textwidth]{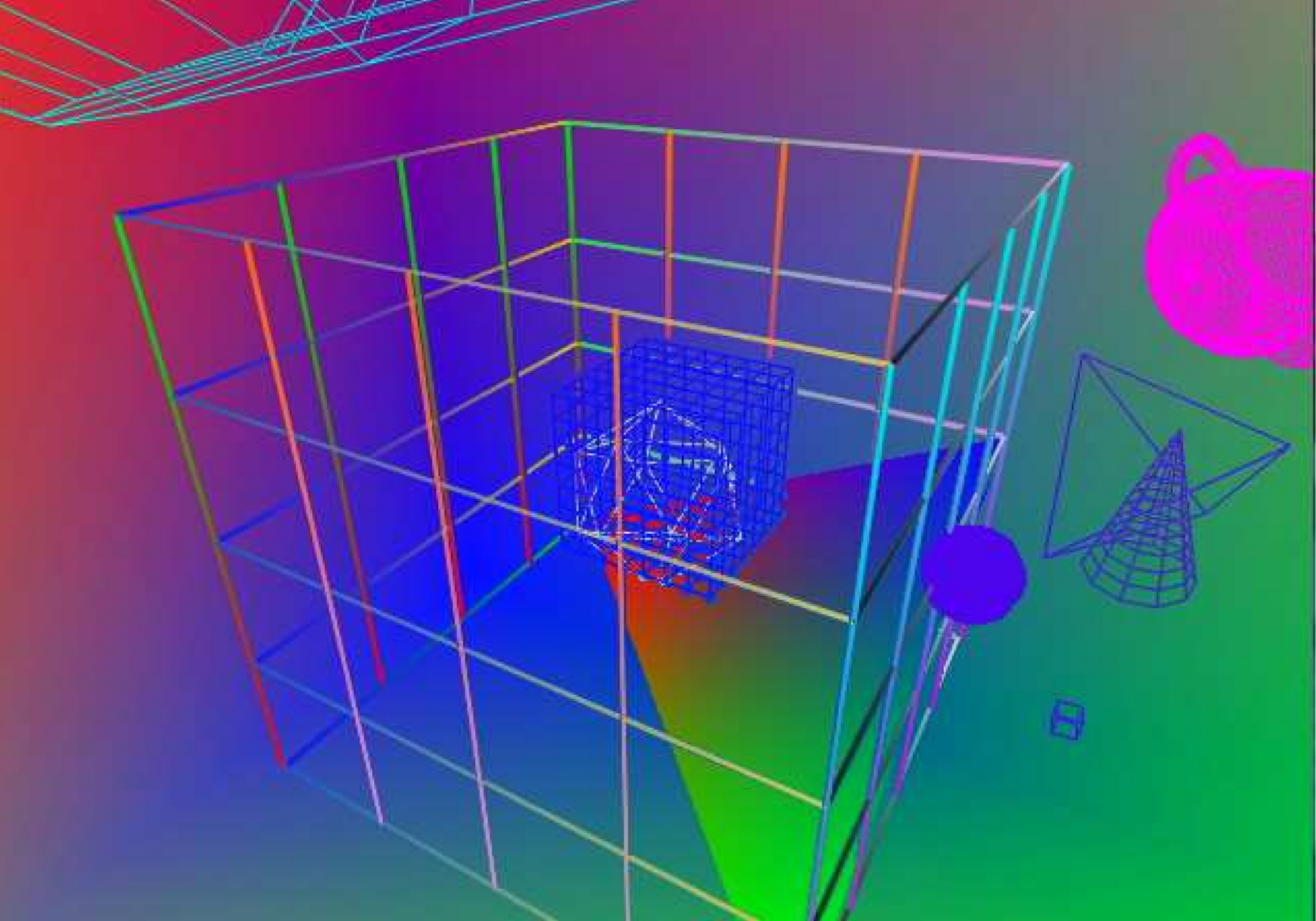}
                                \vspace{-0.5cm}
        \end{subfigure}%
        ~ 
        \begin{subfigure}[b]{0.31\textwidth}
                \includegraphics[width=\textwidth]{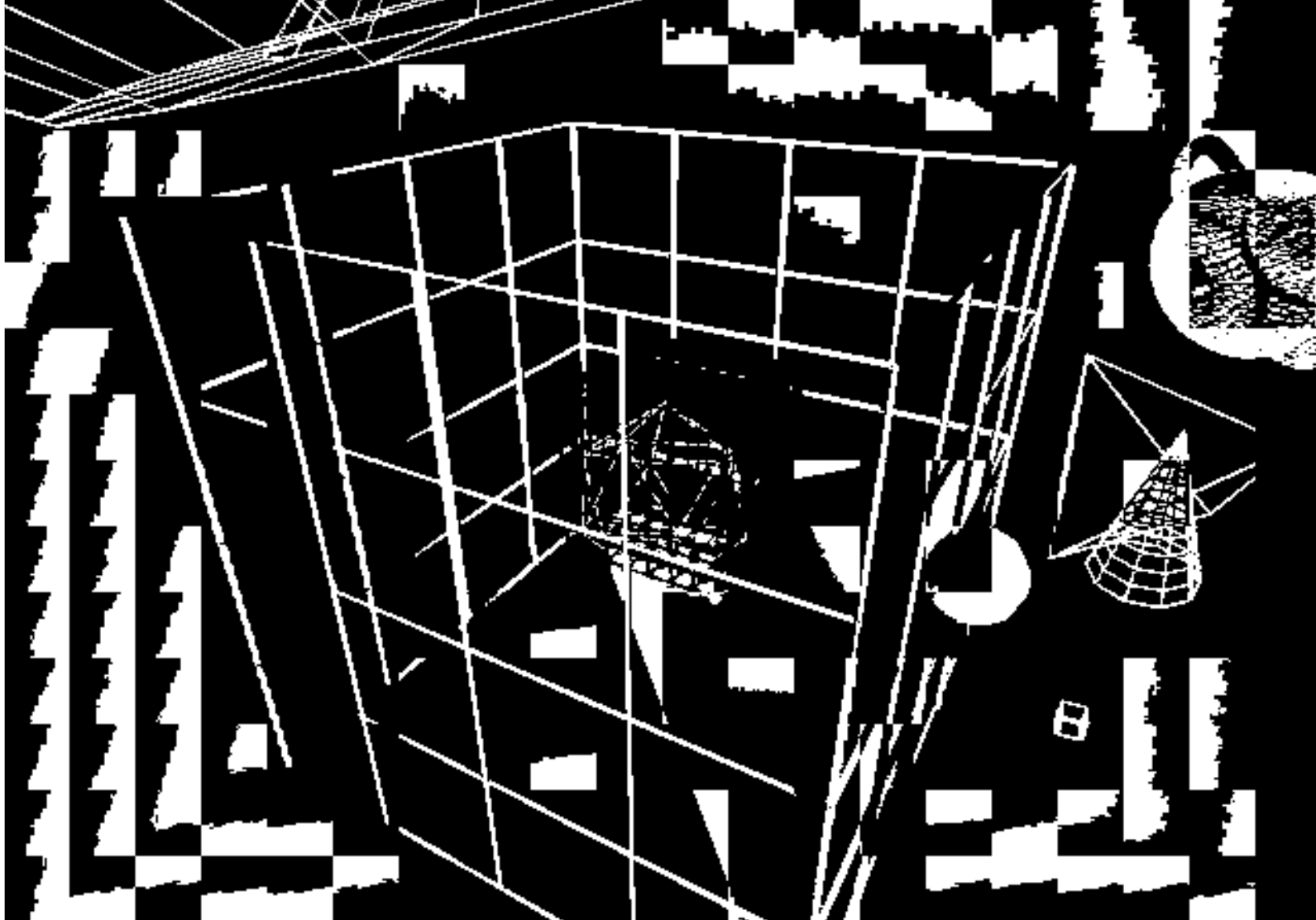}
                \vspace{-0.5cm}
        \end{subfigure}%
        ~ 
        \begin{subfigure}[b]{0.31\textwidth}
                \includegraphics[width=\textwidth]{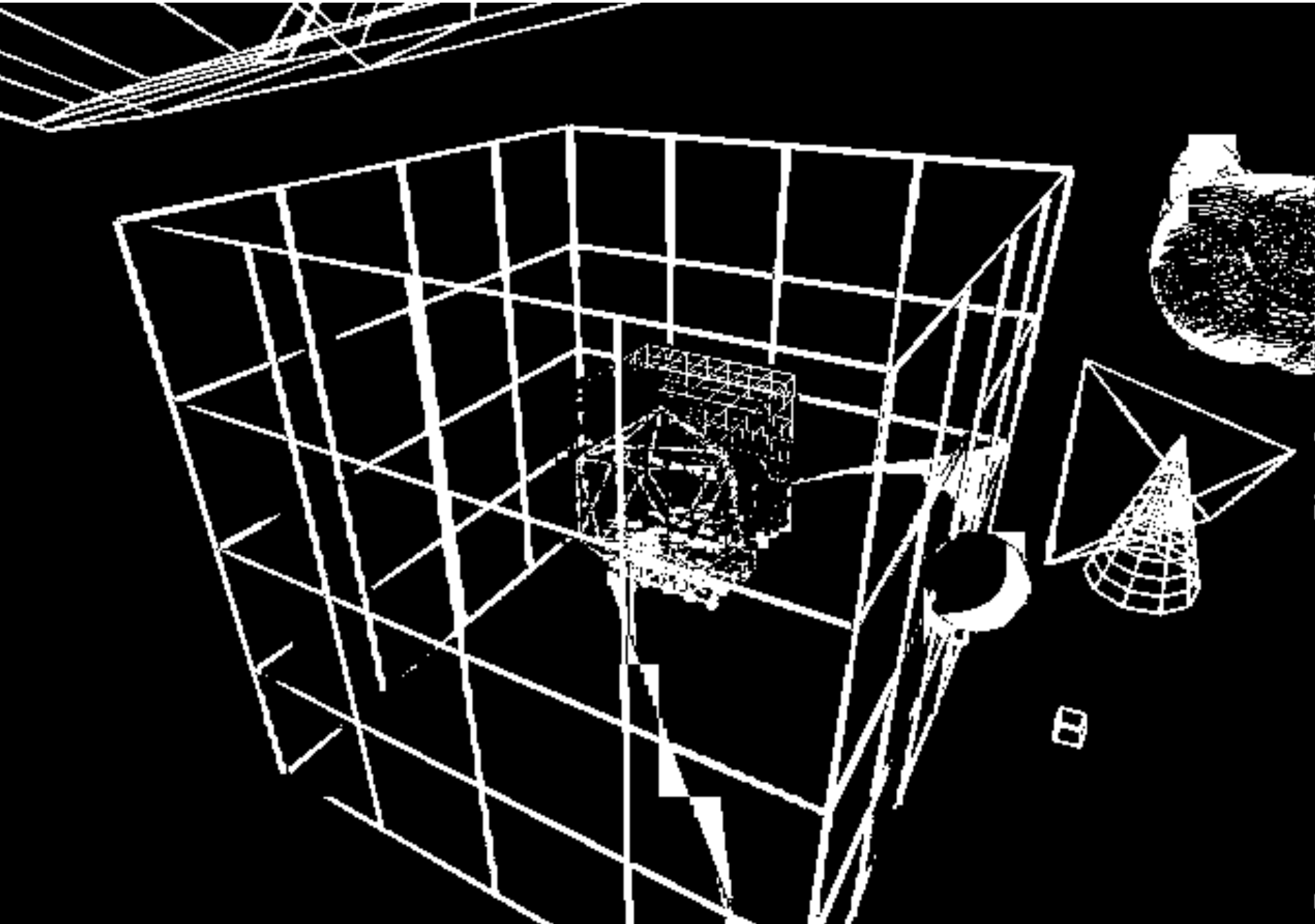}
                 \vspace{-0.5cm}
        \end{subfigure}
        \vspace{0.2cm}
        \caption{Segmentation result for Programming sequence,  the left, middle and right images denote the original, foreground map by hierarchical clustering in DjVu and foreground map by the proposed method respectively}
\end{figure*}

\begin{figure*}
        \centering
       \vspace{-1cm}
        \begin{subfigure}[b]{0.3\textwidth}
                \includegraphics[scale=0.29]{11Original_Image_scissored.pdf}
                                \vspace{-0.5cm}
        \end{subfigure}%
        ~ 
        \begin{subfigure}[b]{0.3\textwidth}
                \includegraphics[scale=0.495]{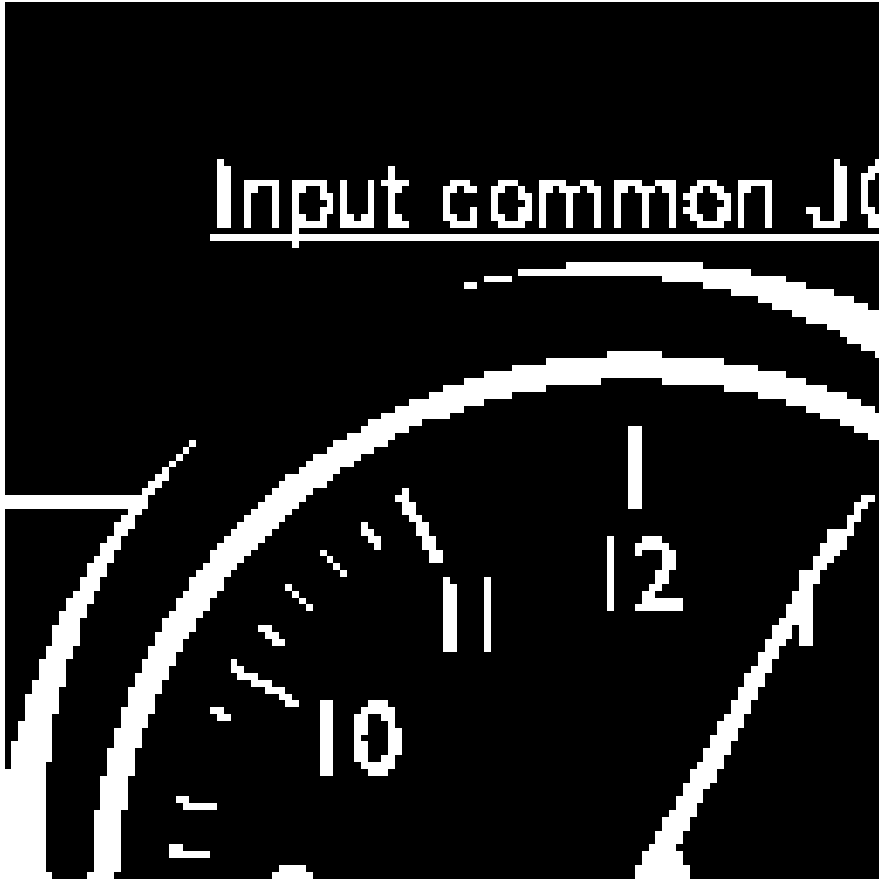}
                \vspace{-0.5cm}
        \end{subfigure}%
        ~ 
        \begin{subfigure}[b]{0.3\textwidth}
                \includegraphics[scale=0.29]{11Foreground_map_scissored.pdf}
                 \vspace{-0.5cm}
        \end{subfigure}
        \vspace{0.6cm}
        \caption{Segmentation result for MissionControl sequence,  the left, middle and right images denote the original, foreground map by hierarchical clustering in DjVu and foreground map by the proposed method respectively}
\end{figure*}

\clearpage
\section{Preliminary Results for Principle Line Extraction in Palmprint Images}
\vspace{-2.95cm}
To demonstrate that the proposed algorithm can be used for other applications, we have applied it for detecting the principle lines from low-resolution palmprint images. Such detection is an important pre-processing step for palmprint recognition \cite{palmprint}.
In general, this task is very challenging due to the low quality of the image and similar colors of background and foreground. Using the proposed scheme followed by a post-processing step to only keep largest connected components, we are able to extract those lines accurately, as shown in Figure 10. For these results, we still used a block size of N=64. But because the background has fine texture, we used more bases to represent the background, with K=14. This number is chosen based on a training set consist of background of palmprint images. The other parameters are the same as the one explained previously. This method can also be applied to fingerprint images, either for extraction of creases or for image enhancement.

\begin{figure}
\begin{center}
    \includegraphics [scale=0.4] {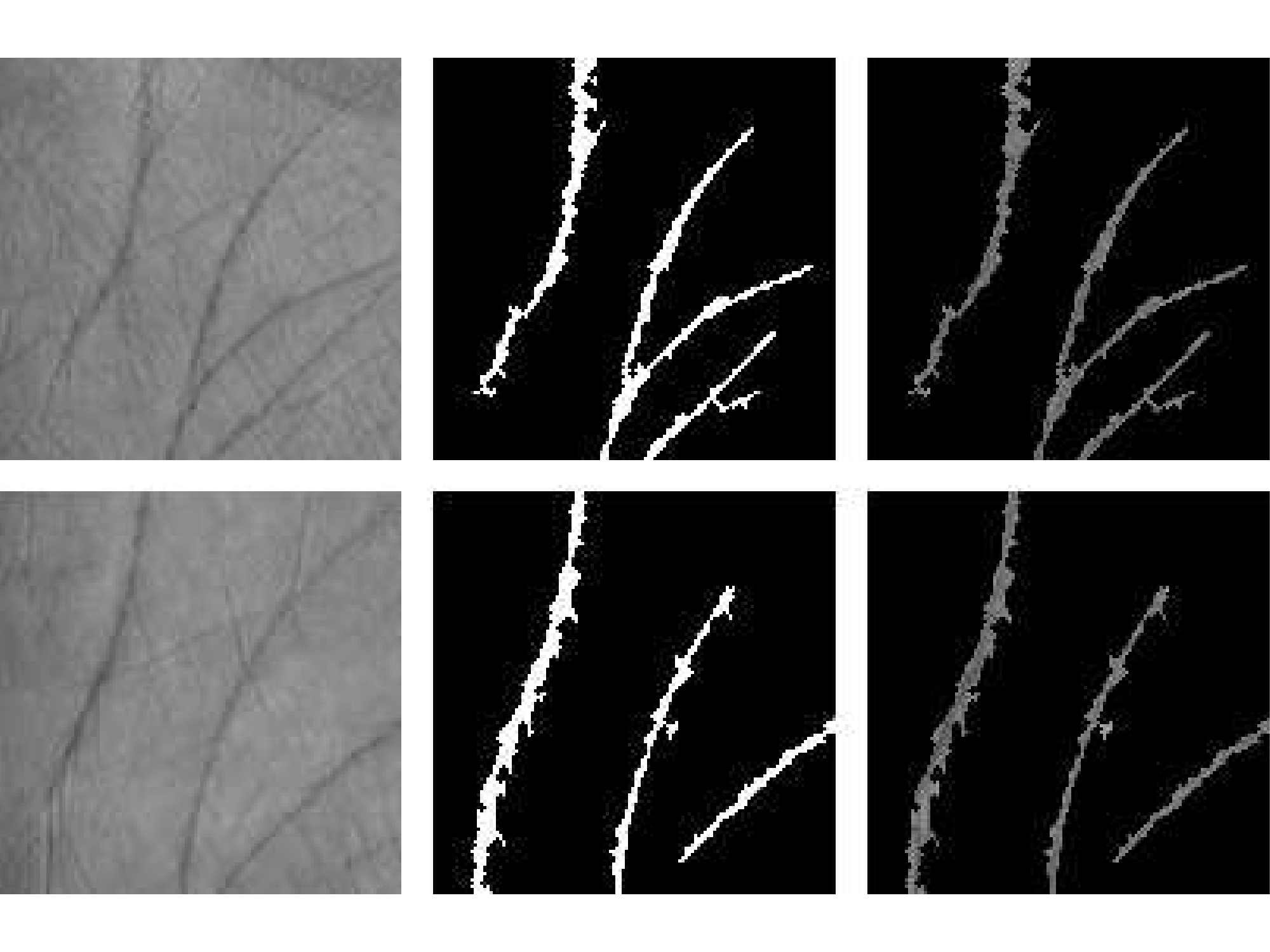}
\end{center}
  \caption{ Principle line extraction from palmprint, the left, middle and right columns denote the original image, foreground map and the foreground image respectively }
\end{figure}

\vspace{-6.5cm}
\section{Conclusion}
\vspace{-4.2cm}
This paper proposed a new algorithm for segmentation of background and foreground. The background is defined as the smooth component of the image which can be well modeled by a set of DCT functions and foreground as those pixels which cannot be modeled with this smooth representation. We propose to use a robust regression algorithm to fit a set of smooth functions to the image and detect the outliers. The outliers are considered as the foreground pixels. RANSAC is used to solve the robust regression problem.
Instead of applying RANSAC to every block, which is computationally demanding, we first check whether the block satisfy several conditions and can  be segmented using simple methods. We further propose to apply the algorithm recursively using quad-tree decomposition, starting with larger block sizes. A block is split only if RANSAC cannot find sufficient inliers in this block. This helps to both improve the segmentation accuracy and to reduce the computation complexity. This algorithm is tested on several test images and compared with one of the well-known algorithms for background/foreground separation and it shows significantly better performance for blocks where the background and foreground pixels have overlapping intensities.
The proposed algorithm is not limited to applications where the background is smooth.  As long as the foreground shows a distinct characteristic from the background, and the background can be represented well with carefully chosen basis functions, then it is possible to separate them using the proposed approach.  We demonstrate this by showing the capability of the proposed algorithm in detecting  principle lines in palmprint images, which have low contrast and complicated texture in the background.

\section*{Appendix. Sparse Decomposition}

Here the sparse decomposition approach for background/foreground segmentation is briefly described. As it is stated earlier, in our segmentation scheme the goal is to represent an image block as a linear combination of $K$ smooth functions, $F=\sum_{k=1}^K \alpha_k P_k$. The foreground pixels cannot be well-represented using this smooth model. Therefore we can add another term, $S$, to this representation such that all pixels in the block can be represented by the new model as $F= \sum_{k=1}^K \alpha_k P_k +S$, where $S$ is a block of the same size as $F$. Now in this representation, the background pixels can be well-represented with the $\sum_{k=1}^K \alpha_k P_k$ term, therefore their corresponding values in $S$ will be zero (or very small). On the other hand, foreground pixels cannot be well represented with the term $\sum_{k=1}^K \alpha_k P_k$. Therefore, their corresponding values in $S$ will be large. So after performing this decomposition any pixel with a large value in its corresponding place in $S$ will belong to the foreground. Here we can use the same idea as RANSAC, where the segmentation is performed by minimizing the number of foreground pixels to minimize the number of non-zero elements in $S$ as:
\begin{gather}
\underset{S,\alpha}{\operatorname{argmin}} \|S \|_0 \\
S.t  \ \ \ \| F-P \alpha-S  \|_2 \leq \epsilon  \nonumber 
\end{gather}
Since $l0$ is not convex, we can replace $l0$ norm with $l1$ to derive the following convex optimization:
\begin{gather}
\underset{S,\alpha}{\operatorname{argmin}} \|S \|_1 \\
S.t  \ \ \ \| F-P \alpha-S  \|_2 \leq \epsilon  \nonumber 
\end{gather}
After solving this problem, $S$ will be an sparse block where any pixel which its corresponding value in $S$ is zero or very small will belong to background.

\end{document}